%% file: acl2023.tex
\pdfoutput=1

\documentclass[11pt]{article}

\usepackage[]{ACL2023}

\usepackage{times}
\usepackage{latexsym}
\usepackage{multirow}
\usepackage{tabularx}
\usepackage{graphicx}
\usepackage{subfigure}
\usepackage{tabularx}
\usepackage{booktabs}
\usepackage{stfloats}
\usepackage{xspace}
\usepackage{amsmath, amssymb}

\usepackage{microtype}
\usepackage{makecell}
\usepackage[T1]{fontenc}

\usepackage[utf8]{inputenc}
\usepackage{microtype}
\usepackage{inconsolata}

\newcommand{\modelname}{\textsc{IAP}\xspace}
\title{Efficient Cross-Lingual Transfer for Chinese Stable Diffusion \\
with Images as Pivots}

\author{
    Jinyi Hu$^{1,2,3}$,\,  Xu Han$^{1,2,3*}$, Xiaoyuan Yi$^{5}$,\, Yutong Chen$^{1,2,3}$, \\
    \textbf{Wenhao Li}$^{1,2,3}$,\, \textbf{Zhiyuan Liu}$^{1,2,3,4,6*}$, \textbf{Maosong Sun}$^{1,2,3,4,6}$\thanks{\ \ Corresponding author.}\\
    $^1$ Department of Computer Science and Technology, Tsinghua University, Beijing \\
    $^2$ Beijing National Research Center for Information Science and Technology \\
    $^3$ IAI, Tsinghua University, Beijing $^4$ IICTUS, Shanghai
    $^5$ Microsoft Research Asia \\
    $^6$ Jiangsu Collaborative Innovation Center for Language Ability, Jiangsu Normal University, Xuzhou \\
    \texttt{hu-jy21@mails.tsinghua.edu.cn}
}

\begin{document}
\maketitle
\begin{abstract}
Diffusion models have made impressive progress in text-to-image synthesis. However, training such large-scale models (e.g. Stable Diffusion), from scratch requires high computational costs and massive high-quality text-image pairs, which becomes unaffordable in other languages. To handle this challenge, we propose \modelname, a simple but effective method to transfer English Stable Diffusion into Chinese. \modelname optimizes only a separate Chinese text encoder with all other parameters fixed to align Chinese semantics space to the English one in CLIP. To achieve this, we innovatively treat images as pivots and minimize the distance of attentive features produced from cross-attention between images and each language respectively. In this way, \modelname establishes connections of Chinese, English and visual semantics in CLIP's embedding space efficiently, advancing the quality of the generated image with direct Chinese prompts. Experimental results show that our method outperforms several strong Chinese diffusion models with only 5\% $\sim$ 10\% training data.
\end{abstract}

\input{introduction}

\input{method}

\input{experiment}

\bibliography{anthology,custom}
\bibliographystyle{acl_natbib}

\appendix

\input{appendix}

\end{document}

%% file: introduction.tex
\section{Introduction}
In recent years, diffusion models \cite{ho2020denoising} have emerged as a promising generative model for various tasks, such as image generation \cite{dhariwal2021diffusion}, speech synthesis \cite{chen2020wavegrad}, molecular generation \cite{xu2021geodiff}, and text-to-image synthesis \cite{ramesh2022hierarchical}. Specifically, large-scale text-to-image diffusion models, such as DALL-E 2 \cite{ramesh2022hierarchical}, Imagen \cite{saharia2022photorealistic}, and Stable Diffusion \cite{rombach2022high}, have gained significant attention for their powerful ability to produce highly realistic and relevant images given a text prompt. 

Despite their promising performance, large-scale diffusion models require massive training resources. For example, Stable Diffusion, the state-of-the-art open-source English text-to-image synthesis model, was trained on billions of text-image pairs. The high resource consumption of training makes it necessary to develop efficient methods for applying these models in various scenarios. Researchers have taken inspiration from the utilization of large-scale pre-trained language models and have designed finetuning methods for few-shot tasks, such as Textual Inversion \cite{gal2022image}, and DreamBooth \cite{ruiz2022dreambooth}. 

To boost the development of text-to-image synthesis around the world, previous works have attempted to train text-to-image models for languages other than English based on Stable Diffusion. In the Chinese community, Taiyi Diffusion \cite{fengshenbang} and AltDiffusion \cite{chen2022altclip} were developed by training a new Chinese text encoder with the main parameters of Stable Diffusion fixed. However, using vanilla training methods, which only optimize the objective function Eq. \eqref{eq:ldm} used in the original training of diffusion model, fails to establish a connection between the new text encoder and the CLIP encoder that provides text embeddings for Stable Diffusion. As a result, even though these models were trained on large datasets, the lack of interaction between the Chinese and CLIP text encoder leads to poor alignment between Chinese, English, and images.

Cross-lingual transfer aims to apply models developed for a language with abundant resources to a relatively low-resource language. Previous studies have utilized the CLIP model \cite{radford2021learning}, a powerful text-image representation model, to learn bilingual or multilingual vision-language models through contrastive objectives \cite{ko2022large, lee-etal-2022-efficient} or knowledge distillation \cite{carlsson-etal-2022-cross}. Unlike simple translation approaches, transferred end-to-end models enable direct alignment of image and target language features, allowing for controllable image generation or editing \cite{hertz2022prompt}. However, achieving cross-lingual transfer on Stable Diffusion remains an unsolved challenge. The main difficulty lies in aligning two sequences of vectors with dynamic sequence lengths, rather than aligning two single pooled vectors. To the best of our knowledge, our work is the first to systematically explore cross-lingual transfer in diffusion-based text-to-image models.


In this work, we propose \modelname\footnote{\ \modelname: \textbf{I}mages \textbf{a}s \textbf{P}ivots}, a simple yet effective approach to transfer Stable Diffusion into Chinese. Similar to the previous work, the main component of Stable Diffusion is kept intact, and only a new Chinese text encoder is trained. Differently, we employ triplets (\textit{image}, \textit{English caption}, \textit{Chinese caption}) as training instances and utilize the image as a pivot to minimize the distance of attentive features between \textit{image} to \textit{English caption} and \textit{image} to \textit{Chinese caption}. This objective promotes the model to learn a representation that is similar to the CLIP text encoder for semantically identical caption pairs. Our experiments demonstrate that our method can achieve superior performance with minimal data compared to various existing Chinese text-to-image models.
\begin{figure*}[t]
\centering
\includegraphics[scale=0.48]{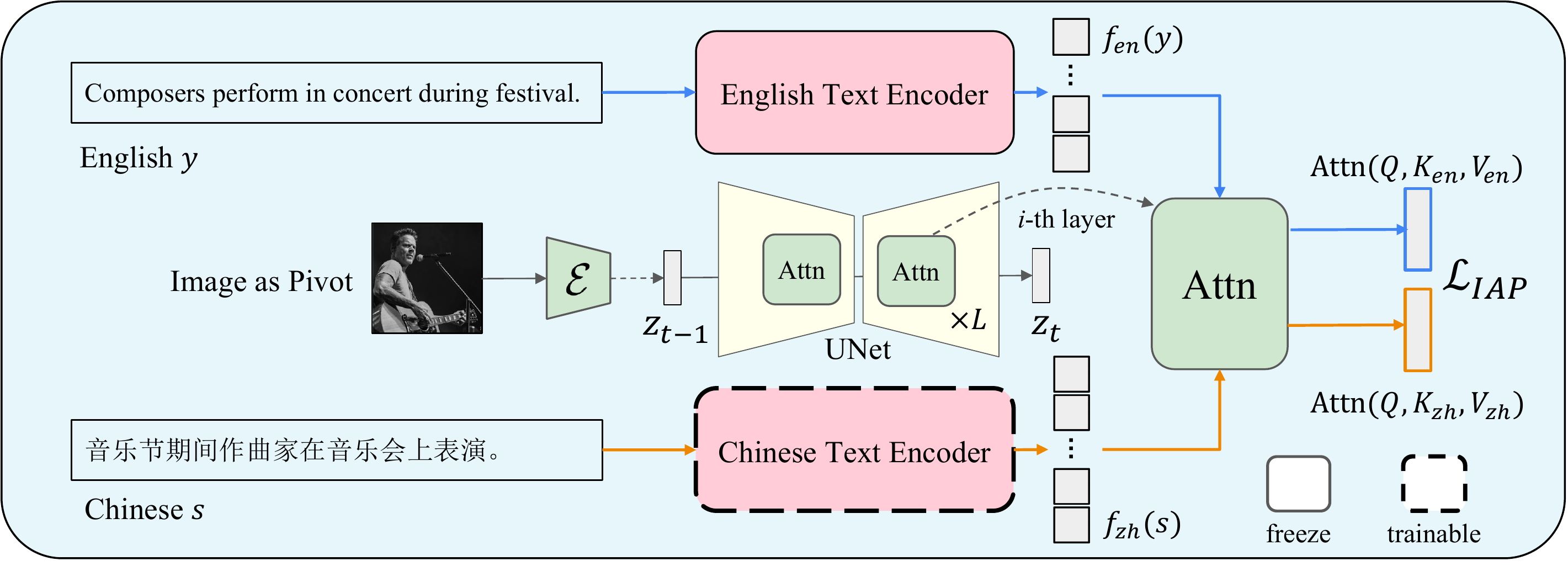} \
\caption{Architecture of \modelname. We fix the parameters of Stable Diffusion and learn a separate Chinese text encoder by considering the images as pivots and minimizing the distance of cross-attention layer output between the images and each language.}
\label{fig:model}
\end{figure*}

\section{Related Work}
\textbf{Cross-Lingual Transfer} Cross-Lingual Transfer has been proven effective in many NLP tasks, such as machine translation \cite{zoph-etal-2016-transfer, ji2020cross}, multilingual pretrained models \cite{conneau-etal-2020-unsupervised} and question answering \cite{lewis-etal-2020-mlqa}. In the case of multi-modal models, some work focused on the cross-lingual transfer of GAN-based text-to-image models \cite{jung-etal-2022-language, zhang2022cje}. Additionally, several works attempted to transfer the powerful vision-language representation model CLIP into other languages \cite{chen2022altclip, carlsson-etal-2022-cross} or benefit multimodal generation \cite{dai-etal-2022-enabling}. However, these methods only involve the final single vector representation, while in Stable Diffusion, we need to align sequences of vectors. To solve this problem, we propose \modelname.

\textbf{Text-to-Image Synthesis} Text-to-image synthesis has been a subject of interest for a long time. In the early stage, GAN was a popular choice as the architecture for text-to-image synthesis models \cite{zhu2019dm, li2019controllable}. With the advent of Transformer \cite{vaswani2017attention}, researchers began utilizing its capabilities for modeling sequences and proposed auto-regressive text-to-image models such as VQGAN \cite{esser2021taming}, Cogview \cite{ding2021cogview}, DALLE \cite{ramesh2021zero} and Parti \cite{yu2022scaling}. Recently, large-scale diffusion models have greatly improved the quality and alignment of generated images, including DALLE-2 \cite{ramesh2022hierarchical}, Imagen \cite{saharia2022photorealistic}, and Stable Diffusion \cite{rombach2022high}.

%% file: method.tex
\section{Method}

\subsection{Preliminaries}
\label{sec:bg}
The architecture of Stable Diffusion is built on the latent diffusion model \cite{rombach2022high} that comprises a CLIP text encoder $f_{en}$, a pretrained image autoencoder $\mathcal{E}$ and $\mathcal{D}$, and a conditional denoising network $\epsilon_\theta(z_t, y, t)$. The pretrained image autoencoder encodes image $x$ into a lower-resolution latent representation $z=\mathcal{E}(x)$ and decodes latent representation $z$ back to image $\hat{x} = \mathcal{D}(z)$. During training, the CLIP text encoder and image autoencoder are frozen, and the conditional denoising network learns a generative process in the latent space conditioned on text prompt representation:
\begin{align}
\label{eq:ldm}
    \mathcal{L}_{LDM} = \mathbb{E}_{\mathcal{E}(x), y, \epsilon\sim\mathcal{N}(0,1),t}||\epsilon_\theta(z_t, f_{en}(y), t) - \epsilon||_2^2,
\end{align}
The specifics of training and sampling for diffusion models can be found in Appendix \ref{sec:deduction}.

The conditional denoising network $\epsilon_\theta(z_t, y, t)$ is essentially implemented by incorporating a cross-attention mechanism \cite{vaswani2017attention} into an underlying UNet architecture \cite{ronneberger2015u}. Formally, at the $i$-th layer, we compute cross-attention as follows:
\begin{align}
\operatorname{Attn}(Q, K, V) =\operatorname{softmax}\left(\frac{QK^{T}}{\sqrt{d}}\right)V,
\end{align}
where the flattened intermediate representation of the image $\varphi_i(z_t) \in \mathbb{R}^{d_\epsilon^i \times h}$ is projected into the query $Q = W_Q^{i}\varphi_i(z_t)$, and the representation of text $f_{en}(y) \in \mathbb{R}^{d_t \times l}$ is projected into key and value $K = W_K^{i} f_{en}(y), V = W_V^{i} f_{en}(y)$. $W_Q^{i} \in \mathbb{R}^{d \times d_\epsilon^i}, W_K^{i} \in \mathbb{R}^{d \times d_t}, W_V^{i} \in \mathbb{R}^{d \times d_t}$ are parameters to be learned.

\begin{table*}[htp]
\centering
\scalebox{0.95}{
\begin{tabular}{cccccccc}
\toprule
\multicolumn{1}{c|}{\multirow{2}{*}{Model}} & \multicolumn{1}{c|}{\multirow{2}{*}{\makecell[c]{Training \\ Data}}} & \multicolumn{2}{c|}{MS-COCO (30K)} & \multicolumn{2}{c|}{COCO-CN (1k)} & \multicolumn{2}{c}{AIC-ICC (30k)}\\ \cline{3-8} 
\multicolumn{1}{c|}{}  & \multicolumn{1}{c|}{} &\multicolumn{1}{c|}{FID$\downarrow$} & \multicolumn{1}{c|}{CLIP$\uparrow$}    & \multicolumn{1}{c|}{FID$\downarrow$} & \multicolumn{1}{c|}{CLIP$\uparrow$}  & \multicolumn{1}{c|}{FID$\downarrow$} & \multicolumn{1}{c}{CLIP$\uparrow$}    \\ \midrule
\multicolumn{1}{c|}{Cogview \cite{ding2021cogview}}  & \multicolumn{1}{c|}{10M}  & \multicolumn{1}{c|}{18.14}   & \multicolumn{1}{c|}{18.2}    & \multicolumn{1}{c|}{-}      & \multicolumn{1}{c|}{-}    & \multicolumn{1}{c|}{-}   &  -  \\ 
\multicolumn{1}{c|}{Cogview2 \cite{ding2022cogview2}}  & \multicolumn{1}{c|}{30M}  & \multicolumn{1}{c|}{24.0}    & \multicolumn{1}{c|}{22.4}    & \multicolumn{1}{c|}{-}      & \multicolumn{1}{c|}{-}    & \multicolumn{1}{c|}{-} &   -  \\ 
\multicolumn{1}{c|}{ERNIE-ViLG \cite{zhang2021ernie}}  & \multicolumn{1}{c|}{145M}  & \multicolumn{1}{c|}{14.70}    & \multicolumn{1}{c|}{22.4}  & \multicolumn{1}{c|}{-}      & \multicolumn{1}{c|}{-}      & \multicolumn{1}{c|}{-}  &   -  \\ \midrule
\multicolumn{1}{c|}{AltDiffusion \cite{chen2022altclip}}  & \multicolumn{1}{c|}{100M}  & \multicolumn{1}{c|}{18.14}    &  \multicolumn{1}{c|}{34.56}      & \multicolumn{1}{c|}{74.04}       &  \multicolumn{1}{c|}{34.67}   &  \multicolumn{1}{c|}{23.07} &  34.09    \\ 
\multicolumn{1}{c|}{Taiyi \cite{fengshenbang}}  & \multicolumn{1}{c|}{120M} &  \multicolumn{1}{c|}{17.09} & \multicolumn{1}{c|}{33.27}  & \multicolumn{1}{c|}{69.64}      &  \multicolumn{1}{c|}{33.54}  &   \multicolumn{1}{c|}{22.51}  &  33.22\\
\multicolumn{1}{c|}{SD + Translation \cite{fengshenbang}}  & \multicolumn{1}{c|}{-} &  \multicolumn{1}{c|}{14.88} & \multicolumn{1}{c|}{\textbf{35.76}}  & \multicolumn{1}{c|}{66.15}  & \multicolumn{1}{c|}{\textbf{36.04}}  & \multicolumn{1}{c|}{\textbf{19.63}} & 34.12 \\
\midrule
\multicolumn{1}{c|}{\modelname}    & \multicolumn{1}{c|}{3M}              & \multicolumn{1}{c|}{\textbf{13.43}} & \multicolumn{1}{c|}{35.35} & \multicolumn{1}{c|}{\textbf{65.43}} & \multicolumn{1}{c|}{35.65}  & \multicolumn{1}{c|}{20.49} & \textbf{34.55} \\ \bottomrule
\end{tabular}}
\caption{Evaluation results for text-to-image synthesis on MS-COCO, COCO-CN, and AIC-ICC datasets. The results for Cogview, Cogview2 and ERNIE-ViLG are copied from their papers. AltDiffusion and Taiyi Diffusion are evaluated with the same setting as our model. SD + Translation means first translating Chinese prompts into English and then feeding them into Stable Diffusion.}
\label{tab:main_task}
\end{table*}

\subsection{Image As Pivots}
\textbf{Problem Formulation} Given a set of triplet training instances $\mathcal{S} = \{(x, y, s)_n\}$, where $x,y,s$ is the image, English caption, and translated Chinese caption, our goal is to learn a new Chinese text encoder $f_{zh}$ while freezing the Stable Diffusion. Ideally, for a Chinese text prompt $s$, the representation $f_{zh}(s) \in \mathbb{R}^{l_s \times d_t}$ is aligned with the representation $f_{en}(y) \in \mathbb{R}^{l_y \times d_t}$. Then, the denoising network can generate a relevant image with $\epsilon_\theta(z_t, f_{zh}(s),t)$. 

As we need to align two sequences of vector $f_{zh}(s) \in \mathbb{R}^{l_s \times d_t}$ and $f_{en}(y_z) \in \mathbb{R}^{l_y \times d_t}$ with different length, commonly used methods, such as contrastive learning or distillation, which are designed for aligning single vector representations at the sentence-level, are not suitable in this scenario.

To solve this problem, we propose \modelname that considers images as pivots for two languages and minimizes the cross-attention output between images and two languages. Formally, given a triplet $(x,y,s)$, we feed two representations $f_{en}(y)$ and $f_{zh}(s)$ into each cross-attention layer of UNet, respectively. As the output shape of the cross-attention layer is identical regardless of sequence length, we can calculate the element-wise mean squared error between the two outputs:
\begin{align}
\label{eq:iap}
    \mathcal{L}_{IAP}\!=\!||\!&\operatorname{Attn}(Q,\!K_{en},\!V_{en})\!-\!\operatorname{Attn}(Q,\!K_{zh},\!V_{zh})||_2^2,
\end{align}
where $Q = W_Q^{i}\varphi_i(z_t), K_{en} = W_K^{i} f_{en}(y), K_{zh} = W_V^{i}f_{zh}(s), V_{en} = W_V^{i} f_{en}(y), V_{zh} = W_V^{i}f_{zh}(s)$.
In this way, the Chinese text encoder is optimized to generate similar keys and values with the CLIP text encoder's to represent images. 
We sum up the mean squared error of all layers of UNet as our final loss objective.

%% file: experiment.tex
\section{Experiment}
\subsection{Implementation Details}
We train \modelname on the CC3M dataset \cite{sharma-etal-2018-conceptual} that consists of around 3 million image-text pairs. The Chinese captions were translated from the original English captions in the dataset using machine translation. Following AltDiffusion and Taiyi Diffusion, we load Stable Diffusion v1-4 checkpoint \footnote{https://huggingface.co/CompVis/stable-diffusion-v1-4}. We initialize the Chinese text encoder with Taiyi Chinese CLIP \footnote{https://huggingface.co/IDEA-CCNL/Taiyi-CLIP-RoBERTa-102M-ViT-L-Chinese}\cite{fengshenbang}, same as Taiyi Diffusion used. More details are listed in Appendix \ref{sec:implementation}. We used the default scheduler PNDMScheduler \cite{liu2022pseudo} with a guidance scale of 7.5 to sample images for 50 steps.

\begin{figure*}[ht]
\centering
\begin{minipage}[]{0.39\textwidth}
\centering
\includegraphics[scale=0.34]{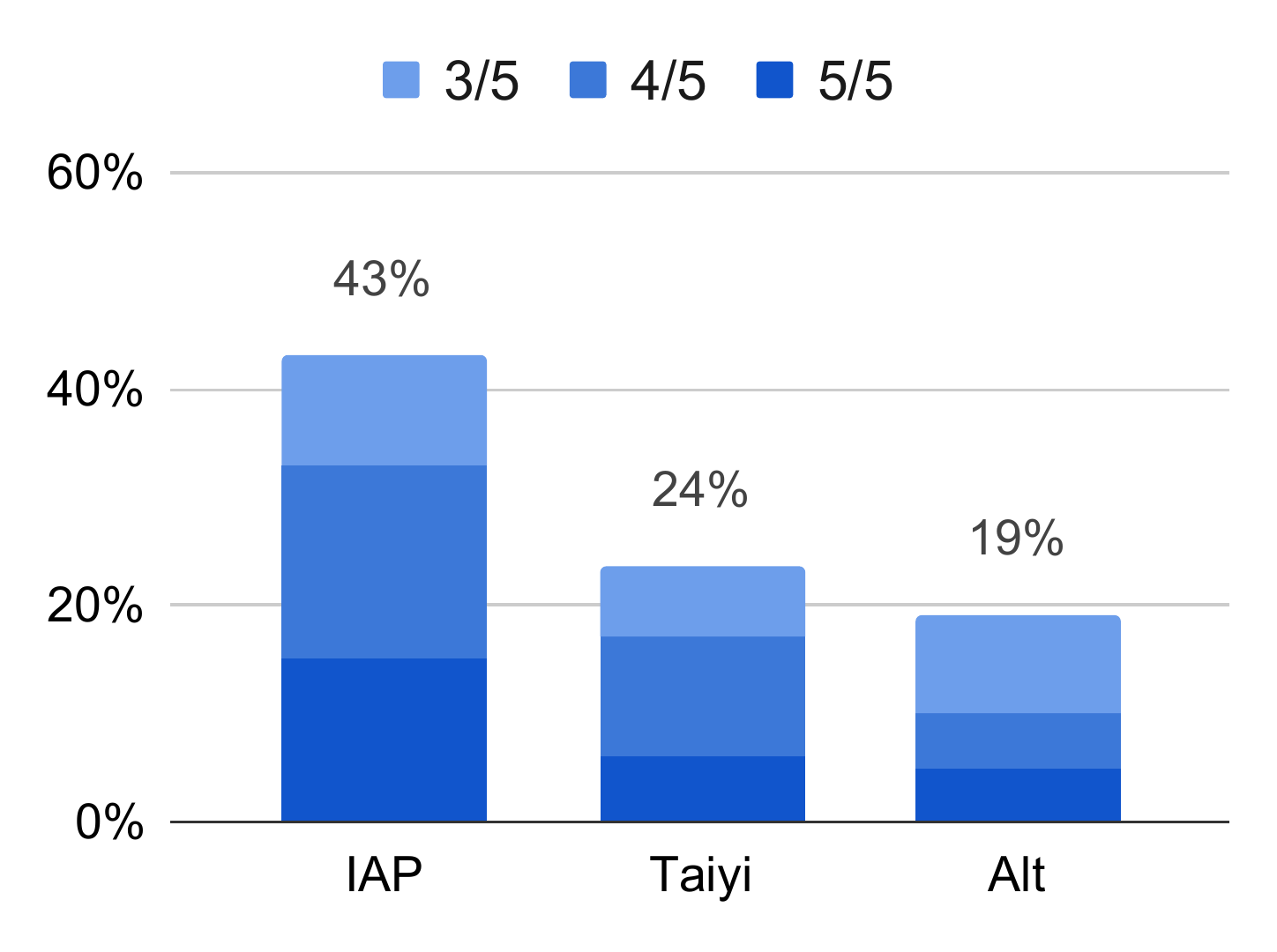}
\includegraphics[scale=0.34]{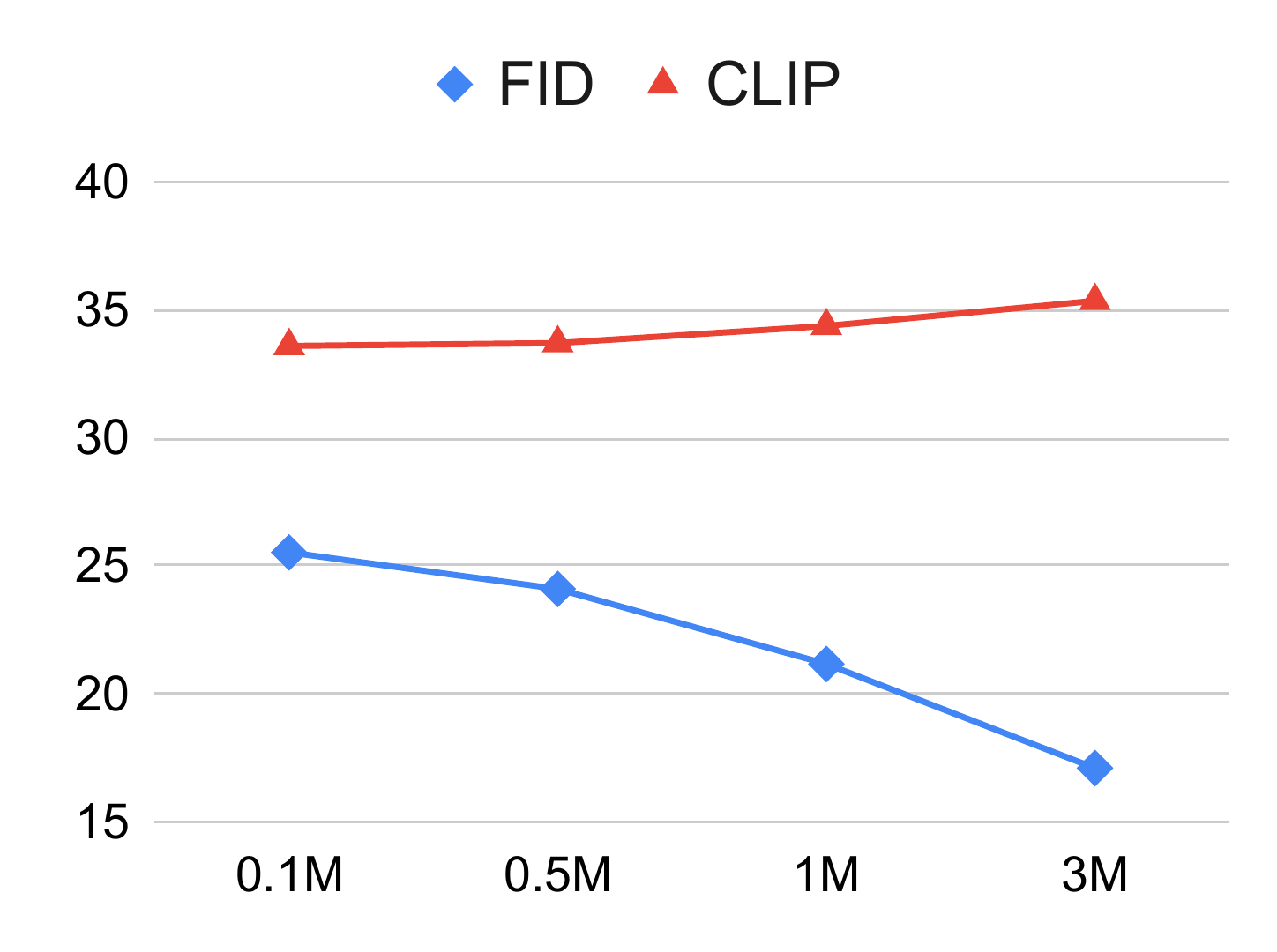}
\caption{The results for human evaluation (top) and \modelname with different sizes of the training dataset (bottom).}
\label{fig:scale}
\end{minipage}
\hfill
\begin{minipage}[]{0.58\textwidth}
\centering
\includegraphics[scale=0.38]{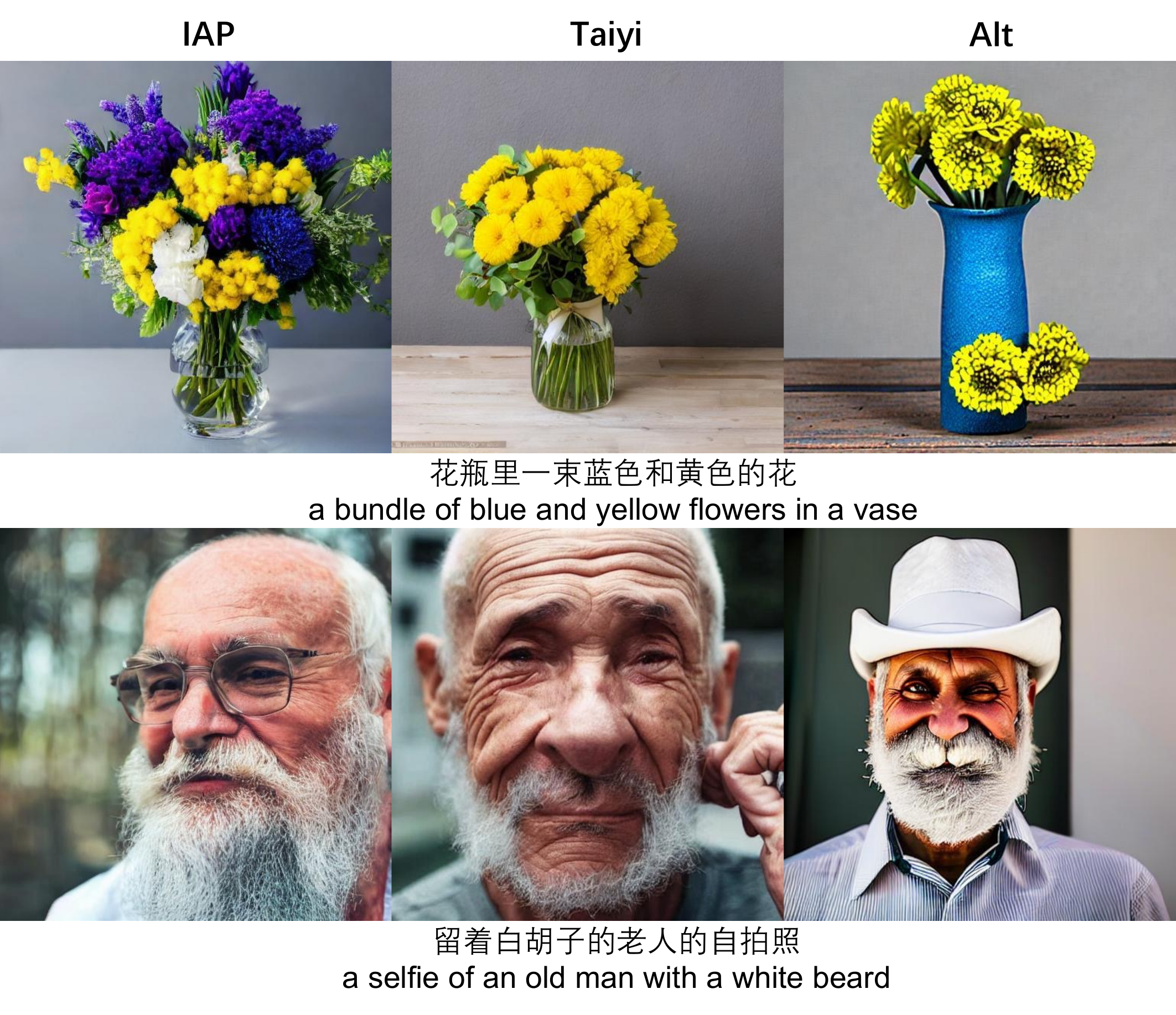}
\caption{Generated images by \modelname, Taiyi Diffusion and AltDiffusion, respectively.}
\label{fig:cases}
\end{minipage}
\end{figure*}

\subsection{Automatic Evaluation}
We evaluate \modelname on the MS-COCO \cite{lin2014microsoft}, COCO-CN \cite{li2019coco} and AIC-ICC dataset \cite{wu2019large} using zero-shot FID score \cite{heusel2017gans}, and CLIP Score \cite{radford2021learning}. FID evaluates the quality and diversity of the generated images, while CLIP Score evaluates the relevance of the generated images to the text. We calculated the FID Score using the \texttt{torch-fidelity} package \cite{obukhov2020torchfidelity} and computed the CLIP Score based on Chinese CLIP \cite{chinese-clip}. The evaluation process is detailed in Appendix \ref{sec:evaluation}. 

The experiment results are presented in Table \ref{tab:main_task}. \modelname achieves satisfactory FID and CLIP scores on all datasets. It outperforms previous methods and performs comparably to a strong translation-pipeline baseline. This demonstrates that \modelname can align Chinese and English semantics explicitly, retaining and transferring the powerful capabilities of Stable Diffusion into Chinese with minimal training resources.


\begin{table}[t]
\centering
\scalebox{0.95}{
\begin{tabular}{c|c|c}
\toprule
Model      &  FID$\downarrow$  & CLIP $\uparrow $\\ \midrule
\multicolumn{3}{c}{Initialization}   \\ \midrule
random              &     18.54     &   32.62           \\
Mengti-T5-encoder   &     13.26     &   34.50      \\
Taiyi Chinese CLIP  &     13.43     &   35.35      \\ \midrule
\multicolumn{3}{c}{Training Loss}   \\ \midrule
Original Loss Eq.\eqref{eq:ldm}      &     104.59      &     27.32         \\
\modelname Loss Eq.\eqref{eq:iap}    &     13.43       &   35.35              \\ \bottomrule
\end{tabular}}
\caption{The results on MS-COCO for models training on CC3M with different initialization for the text encoder and different training loss.}
\label{tab:init}
\end{table}

\subsection{Human Evaluation}
Following \citet{chang2023muse}, we conduct a side-by-side human evaluation to assess the performance of \modelname, AltDiffusion and Taiyi Diffusion. We sample 200 prompts from PartiPrompts \cite{yuscaling}, a curated collection to evaluate model abilities across various categories. We invite five independent annotators and ask them to select the preferable one in each group. More details are presented in Appendix \ref{sec:human}.

As shown at the top of Fig. \ref{fig:scale}, \modelname was selected as the best one in 43\% of input prompts. These results are consistent with the \modelname's superiority on the automatic metrics. 



\subsection{Analysis}
\textbf{Dataset Size} We compare the performance of \modelname training on only partial data pairs with different sizes. As the bottom of Fig. \ref{fig:scale} shows, \modelname achieves satisfactory even with only 0.1M training pairs. Also, with the increasing training data, \modelname consistently improves the FID and CLIP scores, indicating its potential capability for large-scale training.

\textbf{Initialization} We assess the performance of \modelname on MS-COCO using various initialization methods, including random initialization, loading a text-pretrained model (Mengti-T5-encoder \cite{zhang2021mengzi}), and loading a multi-modal pretrained model (Taiyi Chinese CLIP). The results in Table \ref{tab:init} reveal two key findings. Firstly, loading a pretrained model significantly enhances performance. When comparing the multi-modal encoder and the pure text model encoder, Taiyi CLIP slightly outperforms Mengti-T5 encoder, suggesting that pretraining on image-text pairs benefits the text-to-image synthesis task. Secondly, even when trained with random initialization, \modelname still achieves a satisfactory FID score, comparable to AltDiffusion. This indicates the robustness of \modelname to model initialization and its ability to provide incremental improvements with better initialization.

\textbf{Dataset Overlap} To ensure that the improvement in performance on the zero-shot dataset is not due to overlap between the training set (CC3M) and the test set, we trained the text encoder with loss function Eq. \eqref{eq:ldm} on CC3M. As shown in Table \ref{tab:init}, compared with \modelname, vanilla training has very poor performance on the zero-shot testset. This indicates that CC3M does not have a similarity overlap with testset. \modelname shows strong generalization ability in generating images for unseen text prompts.

\subsection{Case Study}
We select two groups of generated images from the human evaluation set. As shown in Fig. \ref{fig:cases}, for the first case, the other two models failed to generate two colors of flowers. For the second case, the details of images from \modelname have the best fidelity. We present more cases in Appendix \ref{sec:apx_case}.
\section{Conclusion}
In this work, we introduce \modelname, an efficient cross-lingual transfer method for the state-of-the-art text-to-image model, Stable Diffusion. By treating images as pivots and minimizing the cross-attentive feature between images and two languages, our new text encoder is able to quickly align with the CLIP text encoder. Despite using fewer data, our model outperforms several Chinese text-to-image models trained with large-scale data. We believe that \modelname can be easily applied to other languages and other diffusion models with similar architectures to Stable Diffusion as well.

\section{Limitations}
While \modelname achieves impressive performance in Chinese text-to-image synthesis, it still has some limitations. First, due to space and resource constraints, we only conducted experiments in Chinese, but we believe \modelname is a general method that can be applied to other languages as well. Second, we observed that the capability for compositional generation is still limited, which is a challenge for all text-to-image synthesis models. We plan to further improve \modelname's ability through structured guidance to tackle this limitation. 

\section{Ethics Statement}
The research and development team behind Stable Diffusion has considered the potential ethical issues that could arise with the use of this model. To ensure that the images generated by text-to-image models are safe for public consumption, they have developed a safety filter to prevent the inclusion of any NSFW information. Similarly, we will follow this process with \modelname to ensure that it is used responsibly and in a healthy manner.

\section*{Acknowledgement}
This work is supported by the National Key R\&D Program of China (No. 2020AAA0106502) and Institute Guo Qiang at Tsinghua University.

%% file: appendix.tex
\section{Detail Deduction of Diffusion Model}
\label{sec:deduction}
Diffusion models \cite{ho2020denoising} are one of the probabilistic generative models that learn the data distribution $p(x)$ by denoising from a reversed Markov Chain constructed by iteratively adding noise to the data:
\begin{align}
    x_t = \sqrt{\alpha_t}x_{t-1} + \sqrt{1 - \alpha_t}\epsilon_t,
\end{align}
The Markov chain $\{x_t\}$ starts with input data $x_0=x$ and end with Gaussian noise $x_T \sim \mathcal{N}(0,1)$. The random noise $\epsilon_t$ is sampled from $\mathcal{N}(0,1)$. Actually, given $x_0=x$, we can directly calculate $x_t$ by:
\begin{align}
\label{eq:x_t}
    x_t = \sqrt{\bar{\alpha}_t}x_{0} + \sqrt{1 - \bar{\alpha}_t}\epsilon,
\end{align}
where $\bar{\alpha}_t=\prod_{i=1}^t{\alpha_i}$. During training, we try to convert the Gaussian noise back into $x_0$. After careful parameterization, diffusion models can be optimized by learning a denoising network $\epsilon_\theta(x_t, t)$ to restore $x_0$ by predicting the noise $\epsilon$:
\begin{align}
    \mathcal{L}_{DM} = \mathbb{E}_{x, \epsilon\sim\mathcal{N}(0,1),t}||\epsilon_\theta(x_t, t) - \epsilon||_2^2,
\end{align}

After training the denoising network $\epsilon_\theta(x_t, t)$, we can make a prediction to $x_0$ at each timestep by replacing $\epsilon$ in Eq. \eqref{eq:x_t} with $\epsilon_\theta(x_t, t)$:
\begin{align}
    x_{0} = \frac{1}{\sqrt{\bar{\alpha}_t}}(x_t - \sqrt{1 - \bar{\alpha}_t}\epsilon),
\end{align}
\begin{align}
\label{eq:x_0t}
    \hat{x}_{0,t}   = \frac{1}{\sqrt{\bar{\alpha}_t}}(x_t - \sqrt{1 - \bar{\alpha}_t}\epsilon_\theta(x_t, t)).
\end{align}
During inference, at timestep $t$, we first predict $\hat{x}_{0,t}$ with Eq. \eqref{eq:x_0t} based on $\hat{x}_t$, then iteratively predict $\hat{x}_{t-1}$ based $\hat{x}_t$ and $\hat{x}_{0,t}$.

For text-to-image synthesis, we need to learn a conditional denoising network $\epsilon_\theta(x_t, y, t)$ and an unconditional denoising network $\epsilon_\theta(x_t, t)$. These two denoising networks can be learned with a single neural network by setting $\epsilon_\theta(x_t, t) = \epsilon_\theta(x_t, y = \emptyset, t)$. During inference, we use classifier-free guidance \cite{ho2022classifier} to sample the image based on the input prompt.

Based on \citet{song2019generative}, the denoising network $\epsilon_\theta(x_t, t)$ is actually predicting the score function:
\begin{align}
    \nabla_x \log q(x) = -\frac{\epsilon_\theta(x_t, t)}{\sqrt{1-\bar{\alpha}}_t}.
\end{align}
Therefore, for the joint distribution $p(x_t,y)$, we have:
\begin{align}
    &\nabla_x \log p(x_t, y) = \nabla_x \log p(x_t) + \nabla_x \log p(y|x_t).
\end{align}
For $\nabla_x \log p(y|x_t)$, we have:
\begin{align}
\begin{split}
    &\nabla_x \log p(y|x_t) \\
                            =& \nabla_x \log p(x_t|y) - \nabla_x \log p(x_t) \\
                           =& -\frac{1}{\sqrt{1-\bar{\alpha}}_t}(\epsilon_\theta(x_t, y, t) - \epsilon_\theta(x_t, t)).
\end{split}
\end{align}
Therefore, we have the classifier-free predictor $\bar{\epsilon}$:
\begin{align}
\begin{split}
    \bar{\epsilon}(x_t, y, t) &= \epsilon(x_t, y, t) + w(\epsilon(x_t, y, t) - \epsilon(x_t, t)) \\
                              &= (w+1)\epsilon(x_t, y, t) - \epsilon(x_t, t).
\end{split}
\end{align}
Here, we introduce the guidance scale $w$ to balance the diversity and alignment of generated images.
\section{Additional Experiment Details.}
\subsection{Implementation Details}
We train \modelname for 100k steps with a batch size of 256 and a learning rate of 5e-5 on 8 NVIDIA A100 GPUs. The entire training process took around 15 hours. We have tested that the training can also successfully run on NVIDIA GeForce RTX 3090. We implemented \modelname with open-source Huggingface Diffusers library \cite{von-platen-etal-2022-diffusers}.

\label{sec:implementation}
\subsection{Automatic Evaluation Details}
\label{sec:evaluation}
Following \citet{ramesh2022hierarchical}, we randomly selected 30,000 images from the MS-COCO validation set and translated the English captions into Chinese using the THUMT open-source toolkit \cite{tan-etal-2020-thumt}. The COCO-CN test set contains 1,000 image-text pairs with manually translated Chinese captions. The AIC-ICC test set also contains 30,000 samples.

For the licenses of the datasets we use, MS-COCO is licensed under a Creative Commons Attribution 4.0 License. COCO-CN and AIC-ICC use MIT License. CC3M is collected by Google.
\subsection{Human Evaluation Details}
\label{sec:human}
We selected 200 prompts from PartiPrompts and translated them into Chinese using machine translation. Then, 200 groups of images are generated by prompting the models with those Chinese inputs. We invited five independent annotators to conduct the human evaluation. During the evaluation process, the annotators were provided with three shuffled images and the input prompt on each page and were asked to assess the alignment between the image and the text, as well as the quality of the images. If more than three annotators chose one image, it was considered the best in this group. We present the results in Fig. \ref{fig:scale}.

\section{Additional Generated Cases}
\label{sec:apx_case}
\begin{figure*}[t]
\centering
\includegraphics[scale=0.5]{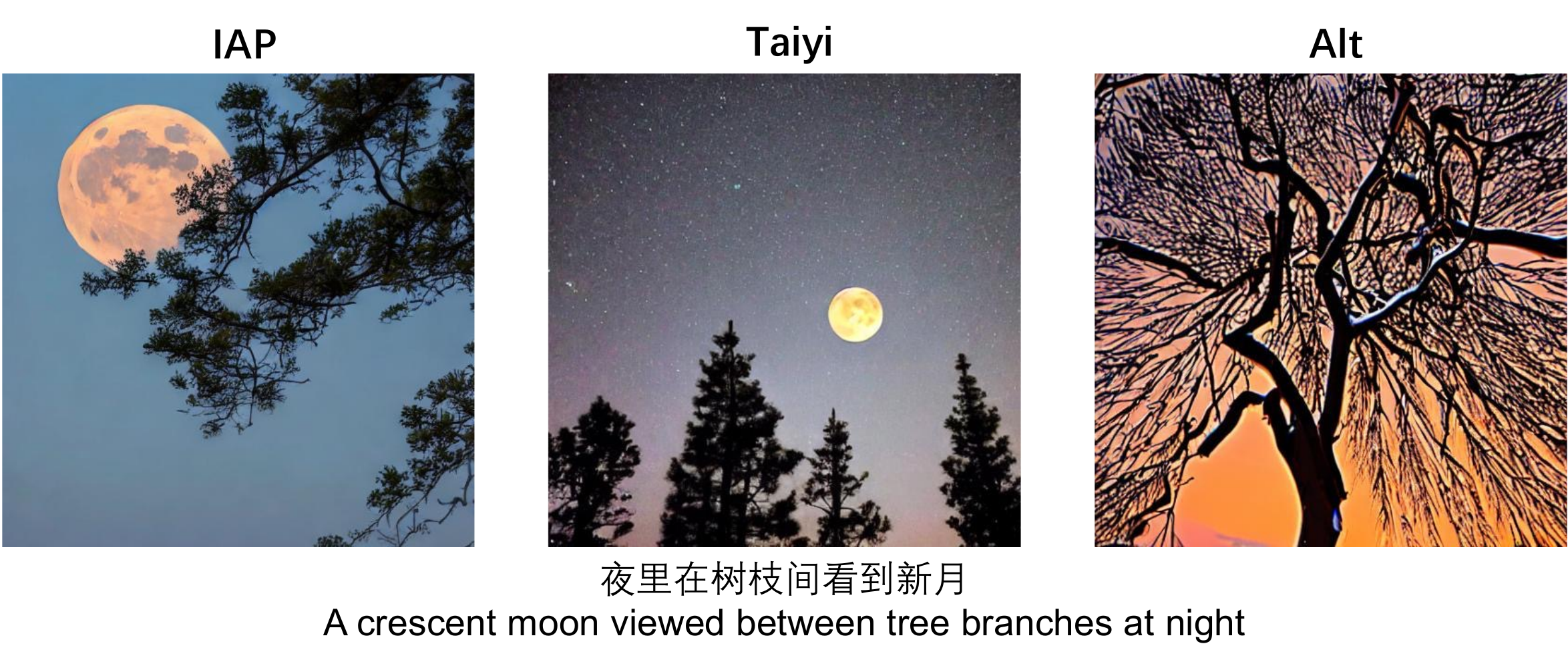}
\includegraphics[scale=0.5]{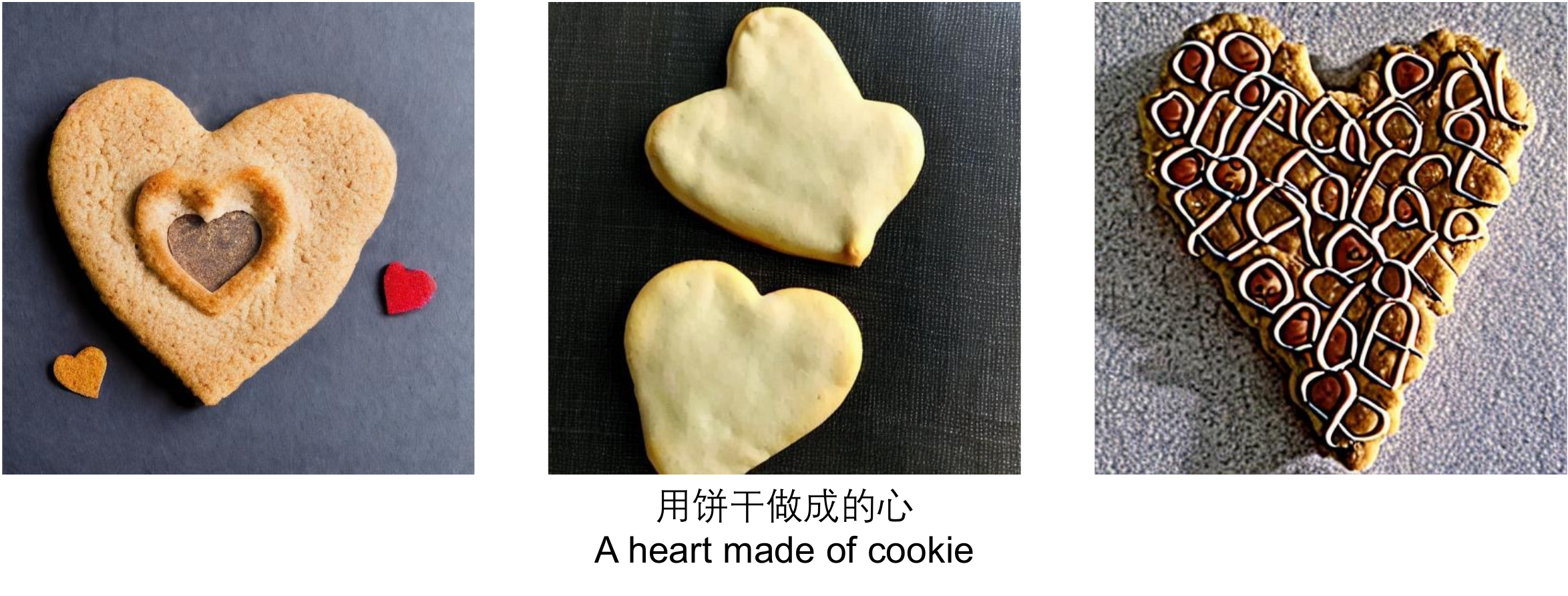}
\includegraphics[scale=0.5]{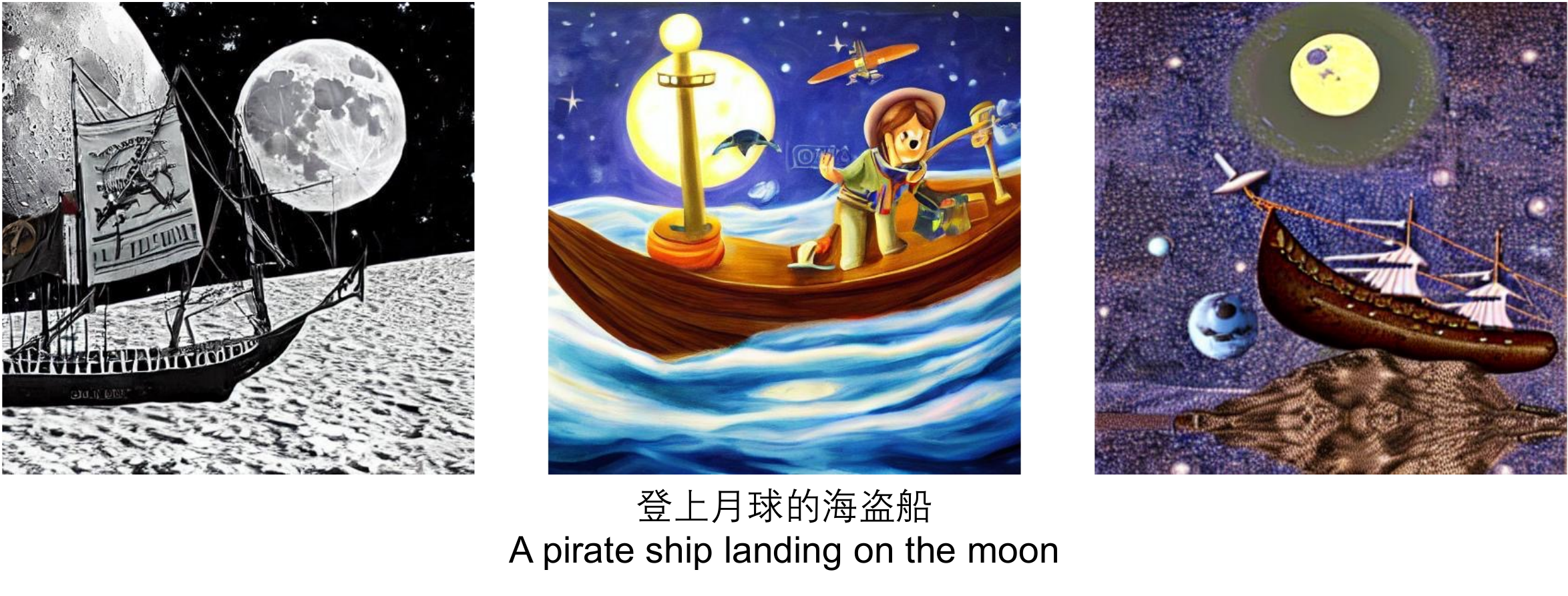}
\includegraphics[scale=0.5]{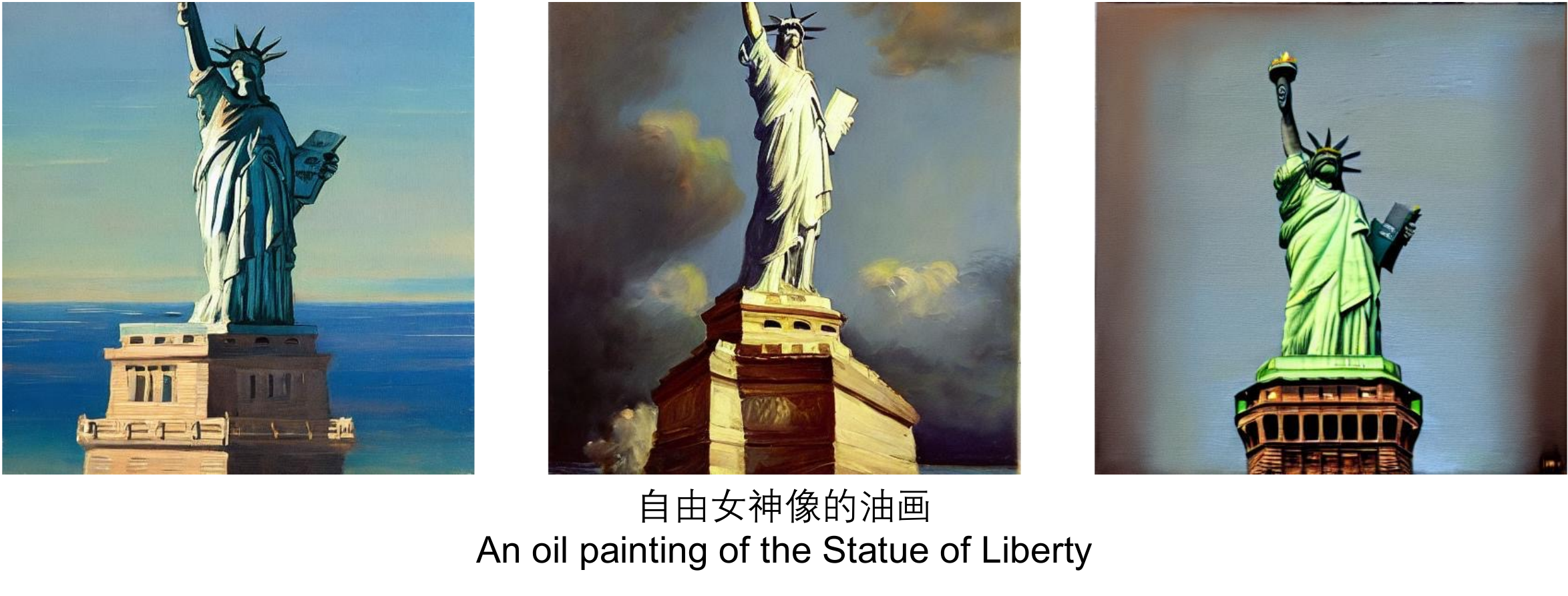}
\includegraphics[scale=0.5]{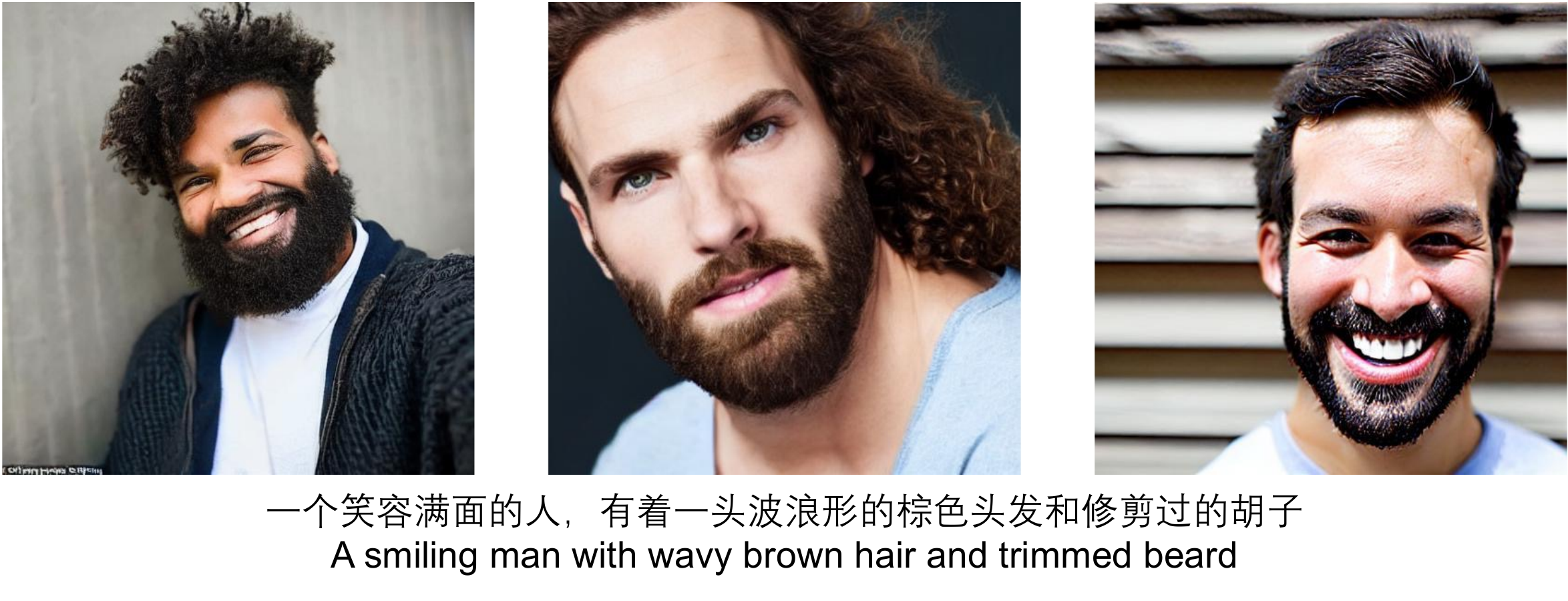}
\caption{Additional generated cases with prompts from PartiPrompts.}
\label{fig:case1}
\end{figure*}

\begin{figure*}[t]
\centering
\includegraphics[scale=0.5]{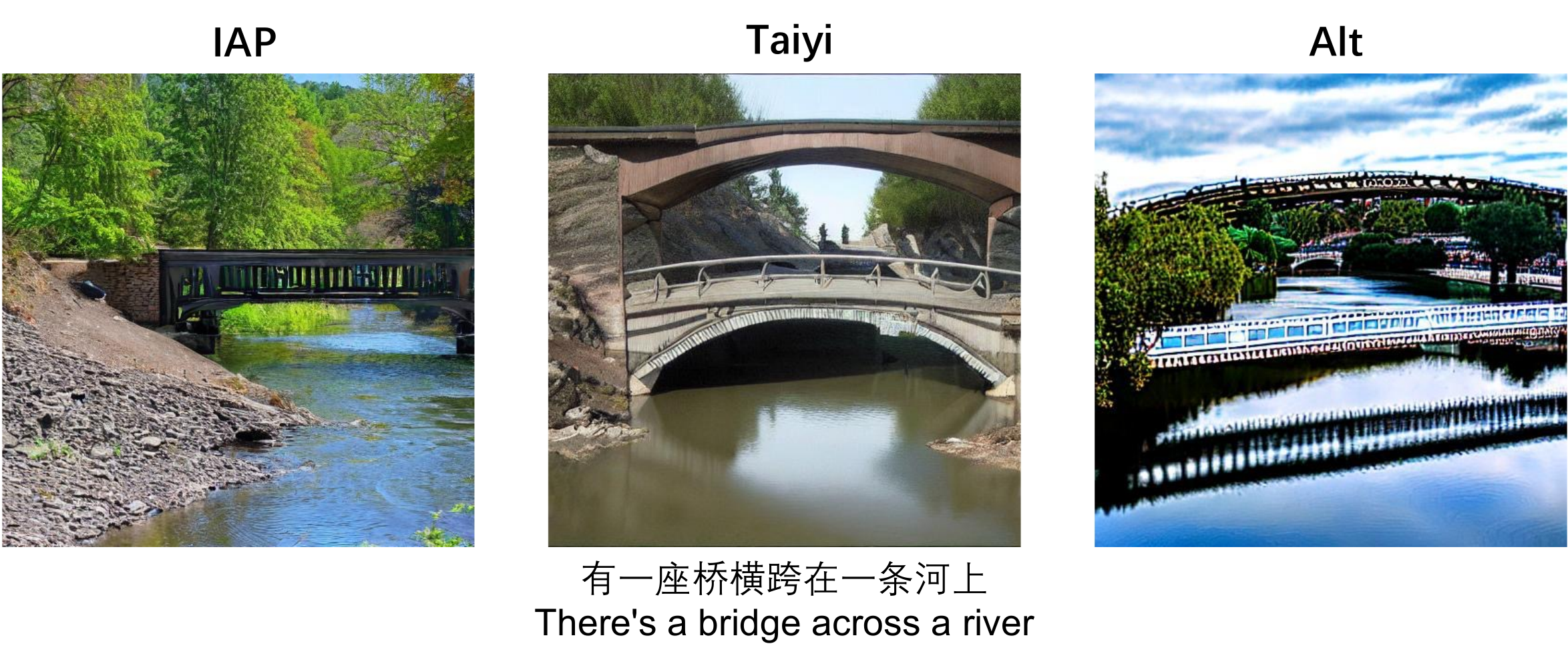}
\includegraphics[scale=0.5]{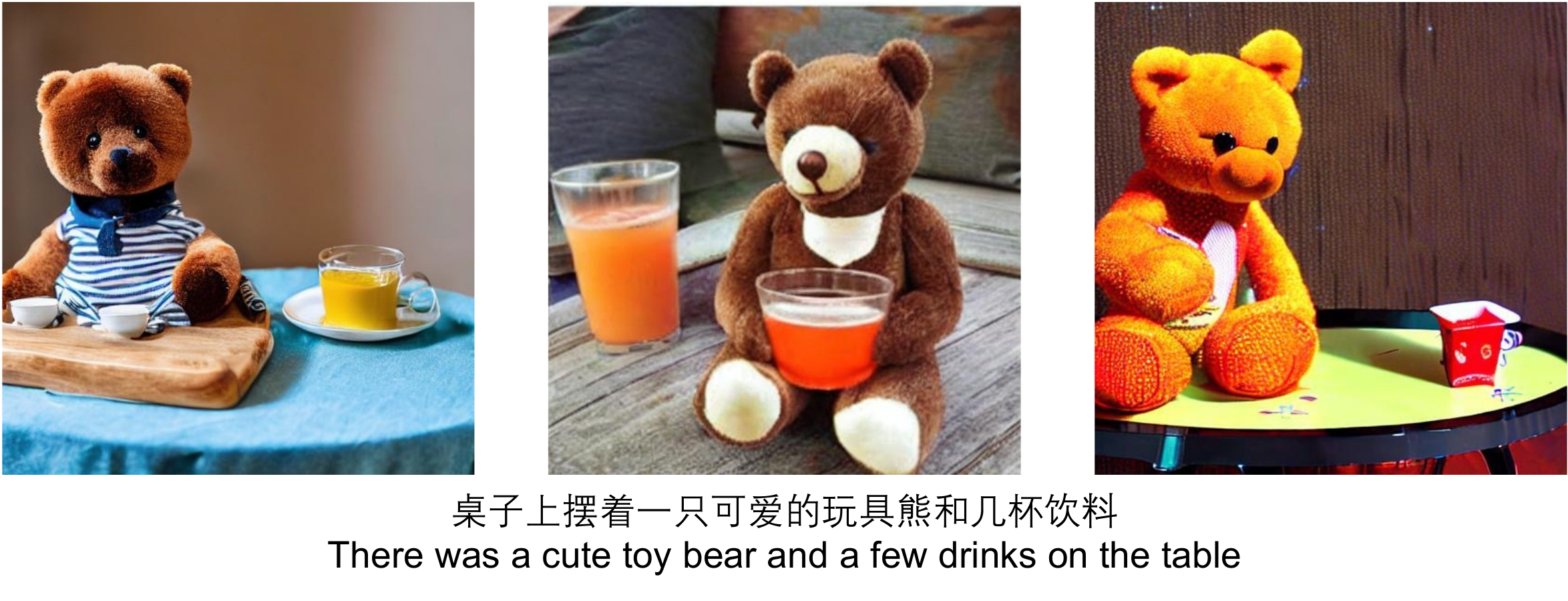}
\includegraphics[scale=0.5]{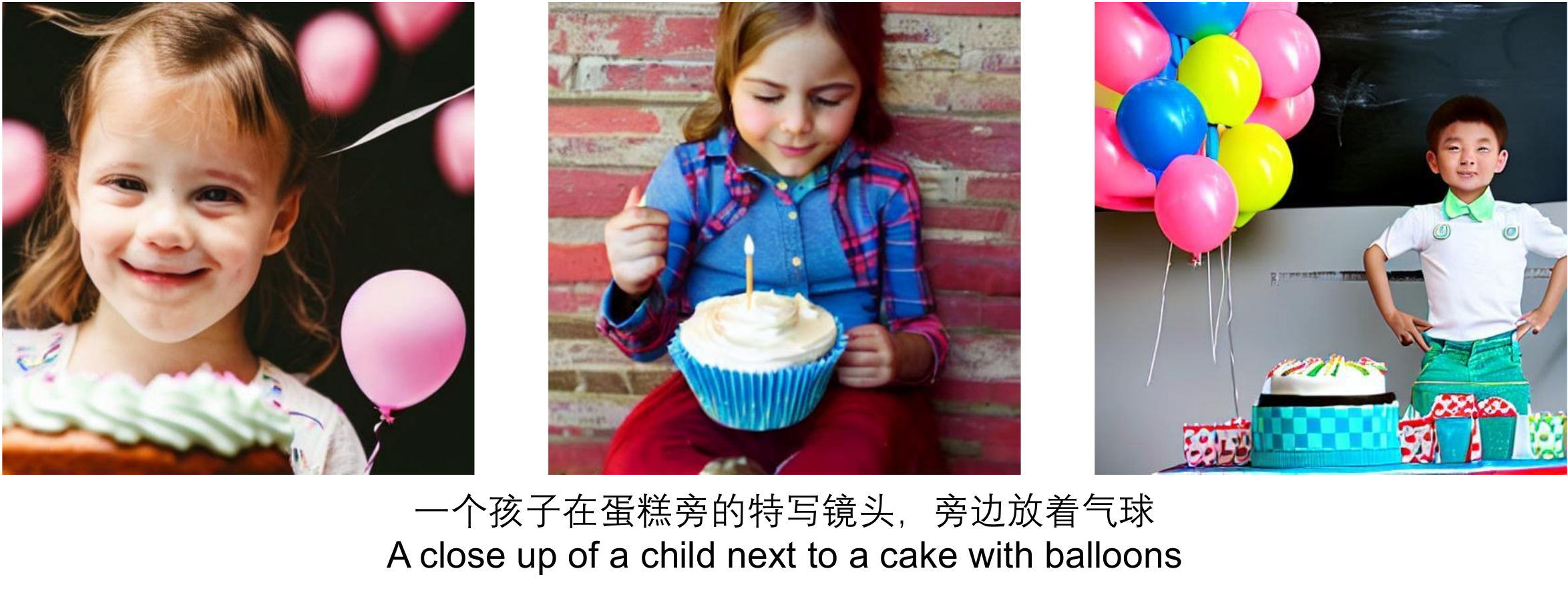}
\includegraphics[scale=0.5]{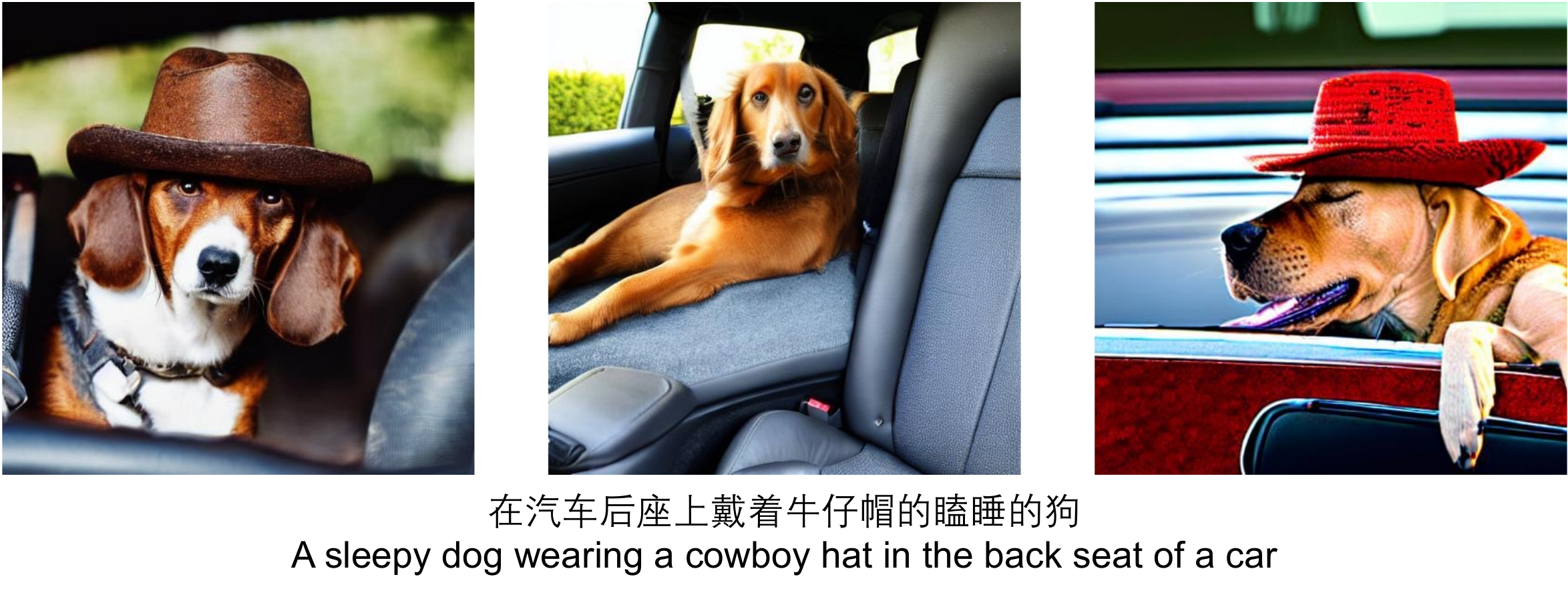}
\includegraphics[scale=0.5]{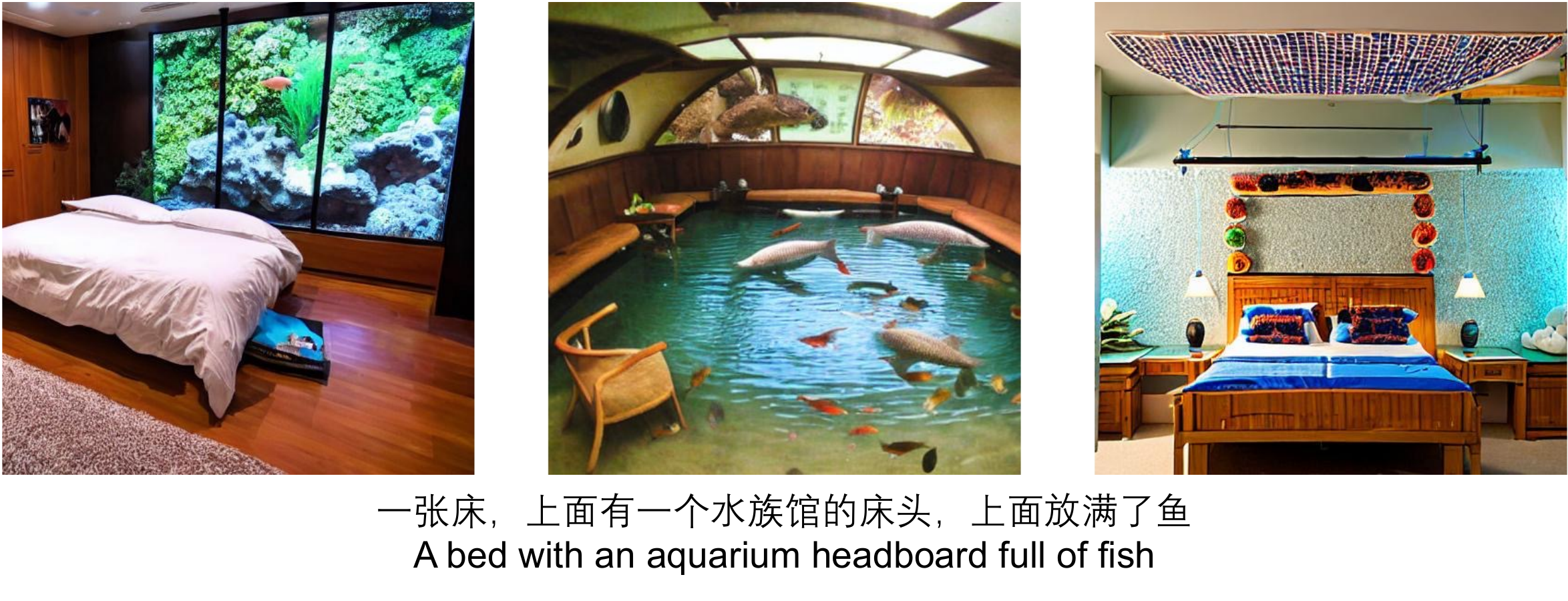}
\caption{Additional generated cases on MS-COCO.}
\label{fig:case2}
\end{figure*}

We present more cases in Fig. \ref{fig:case1} and Fig. \ref{fig:case2}.

